\documentclass{article}

\usepackage{PRIMEarxiv}

\usepackage[utf8]{inputenc} 
\usepackage[T1]{fontenc}    
\usepackage{hyperref}       
\usepackage{url}            
\usepackage{booktabs}       
\usepackage{amsfonts}       
\usepackage{nicefrac}       
\usepackage{microtype}      
\usepackage{lipsum}
\usepackage{fancyhdr}       
\usepackage{graphicx}       
\graphicspath{{media/}}     
\usepackage{tikz}
\usepackage{pifont}
\usepackage{amssymb}

\newcommand{\cmark}{\ding{51}}%
\newcommand{\xmark}{\ding{55}}%
\renewcommand{\P}{\mathcal{P}}
\newcommand{\W}{\mathcal{W}}
\newcommand{\D}{\mathcal{D}}
\newcommand{\A}{\mathcal{A}}
\newcommand{\Name}{\mathcal{N}}
\renewcommand{\u}{u}
\newcommand{\U}{\mathcal{U}}
\newcommand{\TS}{\mathbb{TS}}

\newtheorem{example}{Example}

\newtheorem{definition}{Definition}

\newtheorem{property}{Property}
  
\title{Utility Functions for Human/Robot Interaction}

\author{
  Bruno Yun, Nir Oren \\
  University of Aberdeen \\
  United Kingdom\\
  \texttt{\{bruno.yun, n.oren\}@abdn.ac.uk} \\
   \And
  Madalina Croitoru\\
  University of Montpellier\\
  France\\
  \texttt{croitoru@lirmm.fr}
}

\begin{document}
\maketitle

\begin{abstract}
In this paper, we place ourselves in the context of human robot interaction and address the problem of cognitive robot modelling. More precisely we are investigating properties of a utility-based model that will govern a robot's actions. The novelty of this approach lies in embedding the responsibility of the robot over the state of affairs into the utility model via a utility aggregation function. We describe desiderata for such a function and consider related properties. 
\end{abstract}

\keywords{Human/Robot Interaction \and Utilities \and Aggregation functions}

\section{Should you push babies into lakes?}

Imagine the following scenario. You have bought the latest e-nanny robot that promises to keep your children busy, happy, away from dangerous screens, and to free up time for adults. 
Such a motivating scenario is not far fetched as existing robots, such as Kasper \cite{kasper1,kasper2,kasper3}, have already  shown great success when interacting with autistic children. Another example is the Elias robot-teachers employed, in Finnish schools,  to teach children foreign languages.
Moreover, many robots on the market (e.g. iPal) are now equipped with an ``emotion management system" that is able to detect a child's emotions and respond appropriately \cite{SanoHAN21}.

In our scenario, the robot interacts with children based on the design of an interdisciplinary team of computer scientists, psychologists, neuro-scientists, and roboticists. Such a robot needs to decide how to act and which goals to pursue, and utility-based decision theory provides a well understood approach to implementing such a reasoning system. The robot therefore attempts to maximise its net utility over time. We observe that --- if naively implemented --- such an approach could lead to the robot (for example) pushing a child into a lake in order to find itself in the rewarding scenario where it can save them.  In this paper, we highlight the fact that designing an appropriate utility function for a robot which interacts with humans is (surprisingly) difficult.

Human robot interaction has received much interest from the research community. In the past years, many works in robotics, neurosciences and computer science recognised the need for a human centered interaction.
While most such work focuses on physical joint interaction, i.e. manipulating heavy objects in an industrial setting \cite{BdiwiHWI21,ChenS18} or helping mobility reduced people \cite{mobility,kondo2019frailty} among others, the important topic of joint cognitive interaction has only been getting attention recently \cite{magic-robot,SaundersonN22,Mutlu2016}. 
In this setting the robot needs to collaborate with the human for solving a problem or making a decision. The design of such robot needs to be done with care taking into account aspects such as the trustworthiness of the artificial partner, transparency, accountability and ethical dimension. 

%

The aim of this short paper is twofold. First, we wish to demonstrate that a standard approach to utility, where utilities are associated with specific states, can lead to undesirable behaviour, and that one must instead consider as a property aggregated over multiple states. We then demonstrate that several possible aggregations are unable to capture all possible desirable properties that one may wish them to obtain. We therefore argue that important notions in human-robot interaction, such as responsibility and accountability, are difficult to capture using standard utility-based approaches, and argue that more research in this area is needed.

Let us model our fictitious scenario to show why it poses a problem. By definition, if our robot is utility maximising, it will select those actions which maximise its overall utility. Arguably, saving the child's life nets the robot a very high (positive) utility. Similarly, having the child drown would yield a very negative utility to the robot. Now assume that having the child wet provides a small negative utility. Clearly then, if the robot is confident of its ability to save the child, pushing them into the water and then rescuing them would yield a high positive utility, while not pushing the child in would result in a utility of 0 for the robot. In this paper, we argue that the utility function over sequences of actions should be specified by looking not just at outcomes, but also by considering some notion of blame, accountability and/or responsibility. The core research question we are addressing is how to appropriately model the domain and specify the utility function in order to capture this intuition.


The structure of the paper is as follows.
In Section \ref{sec:background}, we put the basis for the formalisation we will use throughout the paper allowing us to easily capture the notions of possible worlds and actions (transitions between such possible worlds). The reason why we provided our own formalism as opposed to other models, such as situation calculus, was that we wanted to be able to keep the model as light as possible, to focus on the aggregation functions over utilities.
In Section \ref{sec:aggregations}, we  introduce desirable properties that an aggregation function over utilities should satisfy and define two new aggregation functions.
The paper concludes with a discussion on the usefulness of our model in the context of ethical decision making.

\section{Background notions}\label{sec:background}

Let $\P$ be a finite set of propositions having their truth values in $\{ \top, \bot\}$. 
A possible world $w \subseteq \P$ is a subset of propositions where each proposition is true.

\begin{definition}[Possible world]
A possible world in $\P$ is $w \subseteq \P$ such that for every $p \in w$, it holds that $p=\top$. 
\end{definition}

In this paper, we consider the \emph{negation by default}, i.e., the absence of a proposition from a possible world means that it is false. We denote the set of all possible worlds by $\W \subseteq 2^\P$. 
An action is represented by the set of proposition describing the context in which the action can take place and the set of propositions that represent its consequences.

\begin{definition}[Action]
An action $a$ in $\P$ is $(P_a^-,P_a^+)$ where $P_a^-, P_a^+ \subseteq \P$. $P_a^-$ is the \emph{prerequisite} of $a$ and $P_a^+$ is its \emph{consequence}.
\end{definition}

We denote the set of all possible actions by $\A$.
An action $a \in \A$ is applicable to a world $w' \in \W$ iff $P_a^- \subseteq w'$.
Let $w, w' \in \W$, $w$ is \emph{directly accessible} from $w'$ iff there exists an action $a \in \A$ such that $a$ is applicable to $w'$ and $P_a^+ \subseteq w$. 
The closure of a set of worlds by a set of actions is the set of all possible worlds that can be reached by the repeated application of applicable actions.

\begin{definition}[Closure]

If $A \subseteq \A$ and $W \subseteq \W$, the closure of $W$ by $A$, denoted by $W_A$, is the minimal (for set inclusion) subset of $\W$ such that both the following conditions hold:

\begin{itemize}
\item $W \subseteq W_A$
\item if $w \in W_A$ and there exists $a \in A$ such that $w' \in \W$ is directly accessible from $w$ then $w' \in W_A$
\end{itemize}
\end{definition}

We will assume that the sets $\A$ and $\W$ are finite. 

\begin{example}
Let us consider the following example about rescuing babies, where $\P = \{ p_1 , p_2, p_3 \}, \W = \{ w_1, w_2, w_3, w_4 \}$ and $\A = \{ a_1, a_2, a_3 \}$ such that:

\begin{itemize}
\item $p_1$: ``John rescued the baby''
\item $p_2$: ``The baby is in the water''
\item $p_3$: ``John is at home''
\item  $a_1 = ( \{ p_1 \}, \{ \neg p_1, p_2 \})$
\item  $a_2 = ( \{p_2\}, \{ \neg p_2, p_1 \})$
\item  $a_3 = ( \emptyset, \{ p_3 \})$
\item $w_1 = \{ p_1 \}$
\item $w_2 = \{ p_2 \}$
\item $w_3 = \{ p_1 , p_3 \}$
\item $w_4 = \{ p_2 , p_3 \}$
\end{itemize}

The meaning of the actions $a_1, a_2$ and $a_3$ are ``pushing the baby in the water'', ``rescuing the baby'', and ``going home'' respectively.
The world $w_2$ is directly accessible from $w_1$ (this represents the situation where John will push the baby in the water after rescuing it), and $w_3$ is directly accessible from $w_1$ (this represents the situation where John will go home after rescuing the baby). The closure of $\{w_1\}$ by $\A$ is $\{ w_1, w_2, w_3, w_4 \}$. 

\end{example}

Multiple worlds can be accessed from a single world using one action. This allows for external influence and unexpected results. The next example illustrates this intuition.

\begin{example}
Let $\P = \{ p'_1, p'_2, p'_3, p'_4 \}$, $\W = \{ w'_1,$ $w'_2,$ $w'_3 \}$ and $\A = \{ a'= (\{ p'_1 \}, \{ \neg p'_1, p'_2 \}) \}$, where $p'_1$ is ``John has to do his homework'', $p'_2$ is ``John did his homework'', $p'_3$ is ``John obtained a good grade for his homework'' and $p'_4$ is ``John obtained a bad grade for his homework''. If $w'_1 = \{ p'_1 \}$, $w'_2 = \{ p'_2, p'_3 \}$ and $w'_3 = \{ p'_2, p'_4 \}$, it holds that $w'_2$ and $w'_3$ are directly accessed from $w'_1$ using $a'$.
\end{example}

A world $w'$ may also be directly accessible from the world $w$ using multiple actions. This is represented in Example \ref{ex:accessible-different_actions}.	

\begin{example}
Let $\P = \{ p'_1, p'_2, p'_3 \}$, $\W = \{ w, w' \}$ and $\A = \{ a_1= (\{ p'_1 \}, \{ \neg p'_1, p'_2 \}) , a_2= (\{ p'_1 \}, \{ \neg p'_1, p'_3 \}) \}$. If $w = \{ p'_1 \}$, it holds that $w' = \{ p'_2, p'_3 \}$ is directly accessible from $w$ using either $a_1$ or $a_2$.
\label{ex:accessible-different_actions}
\end{example}

In this paper, we assume that such actions can be ``fused'', i.e.\ if there exists $w,w' \in \W, a_1 ,a_2 \in \A$ such that $w$ is directly accessible from $w'$, $a_1$ and $a_2$ are applicable to $w'$ and $P^+_{a_1}, P^+_{a_2} \subseteq w$ then it holds that there exists $a_3 = (P^-_{a_1} \cup P^-_{a_2}, P^+_{a_1} \cup P^+_{a_2})$ in $\A$.
In the next definition, we formalise the notion of maximal actions.

\begin{definition}[Maximal action]
Let $w,w' \in \W$ and $a \in \A$. $a$ is the maximal action from $w$ to $w'$ iff (1) $a$ is applicable to $ w$, (2) $P^+_a \subseteq w'$ and (3) for every $a' \in \A$ such that $a'$ is applicable to $w$ and $P^+_{a'} \subseteq w'$, it holds that $P^-_{a'} \subseteq P^-_{a}$ and $P^+_{a'} \subseteq P^+_{a}$.
\end{definition}

We now assume that we only deal with maximal actions.

\begin{definition}[Transition system]
Let $\W$ be a set of worlds and $\A$ a set of actions in $\P$. A transition system of $\W$ and $\A$ is:
$$\TS = \langle \W,\A, \D , \Name, w_o \rangle$$
Where $w_o \in \W$ and $\D \subseteq \W \times \W$ s.t. $(w,w') \in \D$ iff $w'$ is directly accessible from $w$. The function $\Name: \D \to \A$ returns for any pair $(w,w')$ of $\D$, the maximal action from $w$ to $w'$ in $\A$, and $w_0$ is called the initial state of the transition system.
\end{definition}

\begin{example}
Let us consider $\P = \{ p'_1, p'_2, p'_3 \}$, $\W = \{ w_1 , w_2, w_3 \}$ and $\A = \{ a_1= (\{ p'_1 \}, \{ \neg p'_1, p'_2 \}) , a_2= (\{ p'_1 \}, \{ \neg p'_1, p'_2, p'_3 \}) \}$. If $w_1 = \{ p'_1 \}$, $w_2 = \{ p'_2 \}$ and $w_3 = \{ p'_2, p'_3 \}$, a transition system of $\W$ and $\A$ is $\TS = \langle \W,\A, \D, \Name, w_1 \rangle$ where $\D = \{ (w_1, w_2), (w_1, w_3) \}$ and $\Name((w_1, w_2)) = a_1$. 
A transition system can be represented with a labeled directed graph where the nodes are elements of $\W$, the arcs are elements of $\D$ and the label of an arc from $w$ to $w'$ is $\Name( (w,w'))$. 

%
%
%
%

\end{example}

We now define the notion of sequence of worlds.

\begin{definition}[Sequence of worlds]
Given a transition system $\TS = \langle \W, \A, \D, \Name, w_o \rangle$, a sequence of worlds is $S= [w_0, \dots, w_n]$ such that for every $0 \leq i \leq n-1, (w_i, w_{i+1}) \in \D$.
\end{definition}

The set of all non-empty sequence of worlds in $\TS$ based on $\W$ is $\TS_{\W}$. Note that if $\TS$ has cycles then $\TS_{\W}$ is infinite.
Let $S_1 = [w_0, \dots, w_n]$ and $S_2 = [w'_0, \dots, w'_m]$ be two sequences of worlds, the concatenation of $S_1$ and $S_2$ is $S_1 \oplus S_2 = [w_0, \dots, w_n, w'_0, \dots,$ $w'_m]$.
The size of $S_1$ is denoted by $|S_1| = n$.
The number of occurrences of the world $w$ in a sequence of worlds $S$ is denoted by $occ(w,S)$.
A sequence of maximal actions corresponding to a sequence of worlds $S= [w_0, \dots, w_n]$ in $\TS$ is $S' = [a_0, \dots, a_{n-1}]$ such that for every $0 \leq i \leq n-1, \Name((w_i, w_{i+1}))= a_i$.
There is a unique sequence of maximal actions for any given sequence of worlds. However, multiple sequence of worlds can share the same sequence of maximal actions.
The set of sequences of worlds corresponding to a sequence of maximal actions $S'$ is denoted by $\W_{S'}$. 

From the initial world $w_o$, a robot has to choose between the set of available actions $a$ in $\A$ such that $a$ is applicable to $w_o$ and so on.
The selected action can be determined via a utility function on worlds $\u: \W \to [-1,1]$ which assigns a score to each world.
The set of all possible utility functions on $\W$ is denoted by $\U_\W$.
%
%
In the context of the von Neumann-Morgenstern utility theorem, an robot will always prefer an action that maximises the utility function defined over the worlds \cite{VNMMorgenstern}.
However, a robot will not always choose the immediate preferred action from a world $w$ but plan to maximise the rewards given by the utility function in the long term. 
%

\section{Aggregation over utilities}\label{sec:aggregations}

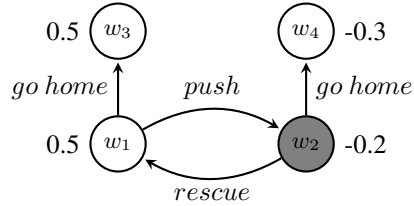
\begin{figure}[!h]
\centering
\begin{tikzpicture}[->,>=stealth,shorten >=1pt,auto,node distance=1.5cm,
 thick,main node/.style={circle,draw,font=\small}]

 \node[main node, label={left:0.5}] (1) {$w_1$};
 \node[main node, label={right:-0.2}, fill=gray] (2) [right of=1, xshift=1cm] {$w_2$};
 \node[main node, label={left:0.5}] (3) [above of=1] {$w_3$};
  \node[main node, label={right:-0.3}] (4) [above of=2] {$w_4$};
    
 \path[->, bend left] (1) edge node  {$push$} (2);
  \path[->, bend left] (2) edge node  {$rescue$} (1);
  
    \path[->] (1) edge node  {$go\ home$} (3);
      \path[->] (2) edge node  [swap] {$go\ home$} (4);
  
 \end{tikzpicture}
 \caption{Representation of a transition system}
 \label{fig-ethical-baby}
\end{figure}

In the real world, it is very complicated to define an robot's utility functions as they usually do not not include the responsibility of the actions that led the robot to the respective world.
Let us illustrate this on the next example formalising the motivating example given in the introduction.
In the world $w_1$, the baby is rescued and thus $\u(w_1)=0.5$. In the world $w_2$, the baby is in the water and thus $\u(w_2)= -0.2$. Notice here that $|\u(w_2)|<| \u(w_1)|$ as the baby might not necessarily drown when in water. 
Clearly, the world where the baby is rescued is better than the world where the baby is in the water.
Thus, it would not be surprising that a robot aiming to maximise its rewards will deliberately push the baby in the water to rescue it afterward. 
This is represented in Figure \ref{fig-ethical-baby}.

In order to capture the responsibility of robots in a transition system $\TS$, we now consider that there is an aggregation function $\sigma: \TS_{\W} \times \U_\W \to [-1, 1]$ that scores how ``good'' a sequence of worlds is w.r.t. a utility function on worlds. 
The set of aggregation functions for a transition system $\TS$ is denoted by $\Sigma_\W$.
Equipped with this utility on sequences of worlds, it is now possible to express that rescuing a baby is better than purposely pushing a baby in the water to rescue it, i.e.\ $\sigma ( [w_2, w_1], u) > \sigma ([ w_2, w_1, w_2, w_1], u )$.
%
%
%
%
The research question we will answer in the reminder of this paper is: ``How do we define $\sigma$ such that it is not advisable for a robot to redo the cycle (rescue, push) because redoing the action does not give them more rewards?''. To this end we will propose two kinds of families of properties: based on cardinality and based on the value of utility function on the possible worlds. We analyse these properties and propose three examples of aggregation functions that respect a subset of such properties.

\subsection{Cardinality properties}

In this section, we investigate cardinality-based properties that an aggregation function over utility values of possible worlds can satisfy.
The first property captures the notion that adding worlds to a sequence of worlds (induced by the subsequent actions) can only decrease its score. The underlying idea is that it is not possible to recover the loss induced by bad actions.

\begin{property}[Non-recovery]
We say that $\sigma \in \Sigma_\W$ satisfies non-recovery iff for every $u \in \U_\W, S_1, S_2, S_3 \in \TS_\W$ such that $S_3 \oplus S_2 \in \TS_\W$ and $\sigma(S_1, u) \geq \sigma(S_3,u)$ then $\sigma(S_1,u) \geq \sigma(S_3 \oplus S_2,u)$.
\label{prop1}
\end{property}

Property \ref{prop2} states that if the score of a sequence of worlds decreased after the addition of some worlds then the added sequence of worlds should have a negative score. 

\begin{property}[Loss conservation]
We say that $\sigma \in \Sigma_\W$ satisfies loss conservation iff for every $u \in \U_\W,S_1, S_2 \in \TS_\W$ such that $S_1 \oplus S_2 \in \TS_\W$ and $\sigma(S_1,u) \geq \sigma(S_1 \oplus S_2,u)$ then $\sigma(S_2,u) \leq 0$. 
\label{prop2}
\end{property}

%

%

Property \ref{prop4} states that repeatedly going through the same sequence of worlds should have a reduced effect on its score and Property \ref{prop5} states that the score of a sequence should not be affected by worlds not in the sequence.

\begin{property}[Redundancy]
We say that $\sigma \in \Sigma_\W$ satisfies redundancy iff for every $u \in \U_\W,S_1, S_2 \in \TS_\W$ such that $S_1 \oplus S_2, S_1 \oplus S_2 \oplus S_2 \in \TS_\W$, it holds that $|\sigma(S_1,u) - \sigma(S_1 \oplus S_2,u)| > | \sigma (S_1 \oplus S_2,u) - \sigma(S_1 \oplus S_2 \oplus S_2,u)|$.
\label{prop4}
\end{property}

\begin{property}[Independence]
We say that $\sigma \in \Sigma_\W$ satisfies independence iff
for every $u \in \U_\W, S_1 = [w_0, \dots, w_n] \in \TS_\W$, $w' \in \W \setminus \{ w_0, \dots, w_n\}$ and $\W' = \W \setminus \{w' \}$, it holds that $\sigma(S_1,u) = \sigma{\restriction}_{\TS_{\W'} \times \U_{\W'}}(S_1,u')$, where $\sigma{\restriction}_{\TS_{\W'} \times \U_{\W'}}$ is the restriction of $\sigma$ to $\TS_{\W'} \times \U_{\W'}$ and $u'$ is the restriction of $\u$ to $\W'$.
\label{prop5}
\end{property}

Property \ref{prop6} states that the benefits gained by adding worlds should be decreasing. Property \ref{prop10} states that the disadvantages obtained by adding worlds should be increasing and, finally, property \ref{prop7} states that the score of the sequence containing only one world should be zero.

\begin{property}[Decreasing benefits]
We say that $\sigma \in \Sigma_\W$ satisfies decreasing benefits iff for every $u \in \U_\W$, $w,w' \in \W$ and $S_1 \in \TS_\W$ such that $S_1 \oplus [w], S_1 \oplus [w] \oplus [w'] \in \TS_\W, \sigma(S_1,u) < \sigma(S_1 \oplus [w],u)$ and $\sigma(S_1 \oplus [w],u) < \sigma(S_1 \oplus [w] \oplus [w'],u)$ then it holds that $\sigma(S_1 \oplus [w] \oplus [w'],u) - \sigma(S_1 \oplus [w],u) < \sigma(S_1 \oplus [w],u) - \sigma (S_1,u)$.
\label{prop6}
\end{property}

\begin{property}[Increasing losses]
We say that $\sigma \in \Sigma_\W$ satisfies increasing losses iff for every $u \in \U_\W, w,w' \in \W$ and $S_1 \in \TS_\W$ such that $S_1 \oplus [w], S_1 \oplus [w] \oplus [w'] \in \TS_\W, \sigma(S_1,u) > \sigma(S_1 \oplus [w],u)$ and $\sigma(S_1 \oplus [w],u) > \sigma(S_1 \oplus [w] \oplus [w'],u)$ then it holds that $\sigma(S_1 \oplus [w] \oplus [w'],u) - \sigma(S_1 \oplus [w],u) > \sigma(S_1 \oplus [w],u) - \sigma (S_1,u)$.
\label{prop10}
\end{property}

\begin{property}[Zero initialisation]
We say that $\sigma \in \Sigma_\W$ satisfies zero initialisation iff for every $u \in \U_\W, w \in \W$, $\sigma([w],u) = 0 $.
\label{prop7}
\end{property}

%

%

\subsection{Value-based properties}

In this section, we investigate value-based properties that an aggregation function can satisfy, i.e.\ how much the score of a sequence of worlds respects a utility function on worlds.

Property \ref{prop12} states that the score of a sequence with only one world should be equal to the utility of that world. 

\begin{property}[Value initialisation]
We say that $\sigma \in \Sigma_\W$ satisfies value initialisation iff for every $u \in \U_\W$ and $w \in \W$, it holds that $\sigma([w],u) = u(w)$.
\label{prop12}
\end{property}

Property \ref{prop13} states that adding a world with a positive (resp. negative) utility to a sequence must increase (resp. decrease) its score.

\begin{property}[Weak additivity]
We say that $\sigma \in \Sigma_\W$ satisfies weak additivity \ref{prop13} w.r.t. $u \in \U_\W$ iff for every $w \in \W$ and $S \in \TS_\W$ such that $u(w) \geq 0$ (resp. $u(w) \leq 0$)
and $S \oplus [w] \in \TS_\W$, it holds that $\sigma(S \oplus [w]) \geq \sigma(S)$
(resp. $\sigma(S \oplus [w]) \leq \sigma(S)$).

\label{prop13}
\end{property}

%
%

Let $u \in \U_\W, S = [w_0, \dots, w_n] \in \TS_\W$ and $K = \{ w \in S \mid u(w) < 0\}$. The aggregation functions that we consider in this paper are defined as follows:
 $$\sigma_{mean}(S,u) = \frac{\sum_{i=0}^n u(w_i)}{n+1}$$

$$
\sigma_{blame}(S,u)  = \left\{
    \begin{array}{ll}
        \sum_{i=1}^n \frac{\min(0,\u(w_i))}{|K|} & \mbox{if } K > 0 \\
        0 & \mbox{otherwise}
    \end{array}
\right.
$$

$$
\sigma_{occ}(S,u)   = \left\{
    \begin{array}{ll}
        \sum_{i=1}^n \frac{\min(0, \u(w_i))}{|K| \times occ(w_i, [w_0, \dots, w_i])} & \mbox{if } K > 0 \\
        0 & \mbox{otherwise}
    \end{array}
\right.
$$


In Table \ref{table:agg}, we show the values returned for the three aggregation functions defined in the paper.
In the Table \ref{table:satisfaction}, we show how the properties proposed in this paper are satisfied by the three aggregation functions defined above.

\begin{table}
\centering
\begin{tabular}{|c|c|c|c|c|}
\hline
Sequence of worlds $S$ & 
$\sigma_{mean}$ & $\sigma_{blame}$ & $\sigma_{occ}$\\
\hline
$[w_2]$ & 
-0.2& 0 & 0\\
$[w_2, w_4]$ & 
-0.25 & -0.15& -0.15\\
$[w_2, w_1]$ & 
0.15 & 0 & 0 \\
$[w_2, w_1, w_3]$ &
0.27 & 0& 0\\
$[w_2, w_1, w_2]$ &
0.03 & -0.1 & -0.05\\
$[w_2, w_1, w_2, w_4]$ &
-0.05 & -0.17 & -0.13\\
$[w_2, w_1, w_2, w_1]$ & 
0.15 & -0.1 & -0.05\\
$[w_2, w_1, w_2, w_1, w_2]$ & 
0.08& -0.13 & -0.06\\
\hline
\end{tabular}
\caption{Aggregated values for sequences on worlds as defined in the transition system of Figure 1.}\label{table:agg}
\end{table}

\begin{table}[h]\label{table1} \caption{Satisfaction of properties of the three illustrative aggregation functions.}
\centering
\begin{tabular}{|c|c|c|c|}
\hline
 & $\sigma_{mean}$  & $\sigma_{blame}$ & $\sigma_{occ}$\\
\hline
Non-recovery & \xmark & \cmark  & \cmark \\
Loss conservation & \xmark & \cmark & \cmark\\
Redundancy & \xmark & \xmark & \cmark \\
Independence & \cmark & \cmark & \cmark \\
Dec. benefits & \xmark & \cmark & \cmark \\
Inc. losses & \xmark & \xmark & \xmark \\
Zero initialisation & \xmark & \cmark & \cmark \\
Value initialisation & \cmark & \xmark & \xmark \\
Weak additivity & \cmark & \cmark & \cmark \\
\hline
\end{tabular}
\label{table:satisfaction}
\end{table}

\section{Discussion}

Standard decision theoretic approaches associate utilities with individual states. As shown above, a naive application of such approaches can lead to undesirable behaviour. Instead, a reasoner seems to require a more complex utility function which aggregates the utility from multiple states in a non-trivial manner to obtain a final utility. However, even in such cases, we are unable to satisfy even relatively simple desirable properties. In other words, we argue that it is far from trivial to define the utility functions that will allow the robots to take the ``best'' course of action. 

In a human-centric setting, an individual would not (for example) push a baby into a pond as they recognise that they will be blamed for the situation, and not rewarded, even if they rescue the child afterwards. These concepts are intimately tied into notions such as responsibility, blame and accountability, and also encapsulate concepts such as causality, blameworthiness, deontic concepts (e.g., permissions, obligations and prohibitions), and --- perhaps most importantly --- intentionality.  In \cite{halpern}  a definition of blameworthiness is provided based on a causal framework. There are other ways to define degree of blameworthiness; for example, through probability with which the harm could have been prevented \cite{FieryCush}, \cite{theoryOfBlame}.

This paper gathers some preliminary ideas, and aims to serve as a ``call-to-arms'' to the community to examine edge cases in utility-based reasoning and ensure that they do not lead to paradoxical or undesirable behaviour. Significant avenues of future work remain open, including the integration of uncertainty into such utility-based systems, and we believe that philosophical work dealing with decision theory \cite{Joyce1999-JOYTFO-4}, as well as work on computational ethics \cite{dennis_see_2020}, can serve to provide additional ideas to deal with the problem highlighted in this paper.

\bibliographystyle{unsrt}  
\bibliography{references}

\begin{thebibliography}{10}

\bibitem{kasper1}
Claire A. G.~J. Huijnen, Monique A.~S. Lexis, and Luc~P. de~Witte.
\newblock Matching robot {KASPAR} to autism spectrum disorder {(ASD)} therapy
  and educational goals.
\newblock {\em Int. J. Soc. Robotics}, 8(4):445--455, 2016.

\bibitem{kasper2}
Sandra Costa, Hagen Lehmann, Kerstin Dautenhahn, Ben Robins, and Filomena~O.
  Soares.
\newblock Using a humanoid robot to elicit body awareness and appropriate
  physical interaction in children with autism.
\newblock {\em Int. J. Soc. Robotics}, 7(2):265--278, 2015.

\bibitem{kasper3}
Joshua Wainer, Ben Robins, Farshid Amirabdollahian, and Kerstin Dautenhahn.
\newblock Using the humanoid robot {KASPAR} to autonomously play triadic games
  and facilitate collaborative play among children with autism.
\newblock {\em {IEEE} Trans. Auton. Ment. Dev.}, 6(3):183--199, 2014.

\bibitem{SanoHAN21}
Taiga Sano, Takato Horii, Kasumi Abe, and Takayuki Nagai.
\newblock Temperament estimation of toddlers from child-robot interaction with
  explainable artificial intelligence.
\newblock {\em Adv. Robotics}, 35(17):1068--1077, 2021.

\bibitem{BdiwiHWI21}
Mohamad Bdiwi, Shuxiao Hou, Lena Winkler, and Steffen Ihlenfeldt.
\newblock Empirical study for measuring the mental states of humans during the
  interaction with heavy-duty industrial robots.
\newblock In Nicolette~M. McGeorge, Alicia Ruvinsky, Mare Teichmann, Leo Motus,
  and Mary Freiman, editors, {\em {IEEE} Conference on Cognitive and
  Computational Aspects of Situation Management, CogSIMA 2021, Tallinn,
  Estonia, May 14-22, 2021}, pages 150--155. {IEEE}, 2021.

\bibitem{ChenS18}
Jen{-}Hao Chen and Kai{-}Tai Song.
\newblock Collision-free motion planning for human-robot collaborative safety
  under cartesian constraint.
\newblock In {\em 2018 {IEEE} International Conference on Robotics and
  Automation, {ICRA} 2018, Brisbane, Australia, May 21-25, 2018}, pages 1--7.
  {IEEE}, 2018.

\bibitem{mobility}
Shunki Itadera, Taisuke Kobayashi, Jun Nakanishi, Tadayoshi Aoyama, and
  Yasuhisa Hasegawa.
\newblock Towards physical interaction-based sequential mobility assistance
  using latent generative model of movement state.
\newblock {\em Adv. Robotics}, 35(1):64--79, 2021.

\bibitem{kondo2019frailty}
Izumi Kondo.
\newblock Frailty in an aging society and the applications of robots.
\newblock {\em Japanese Journal of Comprehensive Rehabilitation Science},
  10:47--49, 2019.

\bibitem{magic-robot}
Jaesik Jeong, JeeHyun Yang, and Jacky Baltes.
\newblock Robot magic show as testbed for humanoid robot interaction.
\newblock {\em Entertain. Comput.}, 40:100456, 2022.

\bibitem{SaundersonN22}
Shane Saunderson and Goldie Nejat.
\newblock Investigating strategies for robot persuasion in social human-robot
  interaction.
\newblock {\em {IEEE} Trans. Cybern.}, 52(1):641--653, 2022.

\bibitem{Mutlu2016}
Bilge Mutlu, Nicholas Roy, and Selma {\v{S}}abanovi{\'{c}}.
\newblock {\em Cognitive Human--Robot Interaction}, pages 1907--1934.
\newblock Springer International Publishing, Cham, 2016.

\bibitem{VNMMorgenstern}
John von Neumann, Oskar Morgenstern, and Ariel Rubinstein.
\newblock {\em Theory of Games and Economic Behavior (60th Anniversary
  Commemorative Edition)}.
\newblock Princeton University Press, 1944.

\bibitem{halpern}
Joseph Halpern and Max Kleiman-Weiner.
\newblock Towards formal definitions of blameworthiness, intention, and moral
  responsibility, 10 2018.

\bibitem{FieryCush}
Fiery Cushman.
\newblock Deconstructing intent to reconstruct morality.
\newblock {\em Current Opinion in Psychology}, 6, 06 2015.

\bibitem{theoryOfBlame}
Bertram Malle, Steve Guglielmo, and Andrew Monroe.
\newblock A theory of blame.
\newblock {\em Psychological Inquiry}, 25:147--186, 04 2014.

\bibitem{Joyce1999-JOYTFO-4}
James~M. Joyce.
\newblock {\em The Foundations of Causal Decision Theory}.
\newblock Cambridge University Press, 1999.

\bibitem{dennis_see_2020}
Louise~A. Dennis.
\newblock Computational goals, values and decision-making.
\newblock {\em Science and Engineering Ethics}, 2020.

\end{thebibliography}

\end{document}